# Sensitivity Analysis for Probability Assessments in Bayesian Networks

**Kathryn Blackmond Laskey**
Department of Systems Engineering and $C^3I$ Center
George Mason University
Fairfax, VA 22030
klaskey@gmu.edu

## Abstract

When eliciting probability models from experts, knowledge engineers may compare the results of the model with expert judgment on test scenarios, then adjust model parameters to bring the behavior of the model more in line with the expert's intuition. This paper presents a methodology for analytic computation of sensitivity values to measure the impact of small changes in a network parameter on a target probability value or distribution. These values can be used to guide knowledge elicitation. They can also be used in a gradient descent algorithm to estimate parameter values that maximize a measure of goodness-of-fit to both local and holistic probability assessments.

## 1 INTRODUCTION

The Bayesian network is becoming a standard representation for uncertainty in symbolic reasoning systems. In a Bayesian network, a probability distribution over a set of random variables is represented by a set of local conditional probability distributions. Local representation has well-known advantages for maintainability of the knowledge base, computation of probabilities, and knowledge elicitation.

To specify a Bayesian network, one first decomposes the hypotheses of interest and related hypotheses into a set of random variables, each representing a set of mutually exclusive and exhaustive hypotheses. Next, one defines a directed acyclic graph which encodes conditional dependencies among the variables. (At this point, auxiliary variables are often added to simplify the conditional dependency structure.) Finally, one specifies the local conditional distributions. At each node, one must define a set of conditional distributions for the node's random variable, one for each combination of values of the direct parents of the node.

The network topology and conditional distributions can be learned from data (e.g., Cooper and Herskovits, 1991; Spiegelhalter and Lauritzen, 1990; Buntine, 1991) or, more commonly in expert systems applications, specified by experts. Although local assessments are sufficient to specify the full joint distribution, these assessments may have non-intuitive consequences for propositions for which beliefs were not assessed directly. Commonly, a Bayesian network model is used to answer queries about the conditional distribution of a *target* variable given specific values for a subset of variables called *evidence* variables. (Sometimes the variables designated as target and evidence variables are fixed in advance; at other times they vary from query to query.) It is rarely the case that the first attempt at assessing network structure and local conditional distributions leaves the expert satisfied with the system's responses to test queries. Knowledge elicitation thus involves an iterative process of specifying a Bayesian network, processing a representative set of queries, and adjusting structure and parameters until the expert is satisfied with the system's response to the test queries.

Key to this iterative process is sensitivity analysis, or examination of the impact of changes in base assumptions on model results. One common approach to sensitivity analysis is to define reasonable ranges for each of the model parameters, vary each parameter from its lowest to highest reasonable value while holding the other variables fixed, and examine the resultant changes in the target value. Another approach, the one taken here, is to compute the partial derivative of the target value with respect to each of the model parameters. A disadvantage of the partial derivative approach is that it provides information only about changes in a small neighborhood about the assessed value. It has the advantage that the expert need not assess upper and lower reasonable values for each parameter. Another advantage for the Bayesian network application is that standard belief network propagation algorithms are easily modified to compute sensitivity values. I show below that the partial derivative method requires far fewer iterations through the belief propagation algorithm than does the direct variation method. Partial derivatives and direct variation can be used in conjunction. Partial derivatives may be used to identify variables with potentially high impact. Reasonable ranges may then be assessed for these variables and sensitivity values may be computed by direct variation.

This paper presents a method for computing sensitivity values. I begin with a single scenario, in which the probability distribution is considered for a target variable given values for a set of evidence variables. These sensitivities may be used to provide feedback to guide the expert in deciding which local probability assessments to consider changing when a global probability estimate is unsatisfactory. Sensitivity values can also be used to suggest possible structural changes to the network. The methodology can also be used to aggregate assessments for multiple scenarios into a best-fit model.

## 2 COMPUTING SENSITIVITY VALUES

### 2.1 Definitions and Notation

Consider a Bayesian network model for $X$, a random vector with $n$ components.[1] Let $X_i$ denote the $i$th component of $X$, and let $X_{c(i)}$ represent the set of conditioning variables, or direct parents, of $X_i$. The probability distribution for $X$ depends on a parameter vector $\theta$. The probability that $X$ takes on a particular value $x$ is given by

$$P(x|\theta) = \prod_i P(x_i|x_{c(i)}, \theta_i) \qquad (1)$$

The factors in (1) are the local conditional probabilities that $X_i = x_i$ given the values $X_{c(i)} = x_{c(i)}$ of the conditioning variables.

The local conditional distribution for $X_i$ consists of a probability for each possible value of $X_i$ given each combination of possible values of its parent variables $X_{c(i)}$. A number of authors have discussed models for the local conditional distributions that depend on fewer parameters. The parameter vector $\theta_i$ has been introduced to permit the methodology to handle such models.

The local parameter vector $\theta_i$ determines the probability distribution of the $i$th node given values of its parent nodes. That is, the parameter vector $\theta$ is partitioned as $\theta = (\theta_1, \ldots \theta_n)$, where $\theta_i$ controls the distribution of $X_i$ given its parents $X_{c(i)}$. The components of $\theta_i$ are denoted by $\theta_{i[k]}$, where $k$ indexes the elements of the parameter vector.

Example 1 below considers the simplest model for the local conditional distributions, one with no restrictions on the local conditional probabilities except that the probabilities in each local conditional distribution sum to 1. Example 1 shows one way to parameterize this model.

*Example 1: Unrestricted Local Conditional Distributions*. The parameter vector $\theta_i$ consists of one component for each combination of values of $X_i$ and its parents. Components of $\theta_i$ are denoted $\theta_{i[x_i, x_{c(i)}]}$ and are restricted to be nonnegative. The conditional probability distributions are related to the parameters as follows.

$$P(x_i|x_{c(i)}, \theta_i) = \frac{\theta_{i[x_i, x_{c(i)}]}}{\sum_{x'_i} \theta_{i[x'_i, x_{c(i)}]}} \qquad (2)$$

(For the initial assessed model it is convenient to scale the $\theta$'s so that the sum in the denominator is equal to 1. However, it is necessary to write the local distribution as (2) in order to enforce the sum-to-one constraint when a single $\theta_{i[k]}$ is changed).

The noisy-OR model (Pearl, 1988) is a commonly cited example of a model for local conditional distributions that depends on fewer parameters than the total number of local conditional probabilities. A noisy-OR model applies when $X_i$ and its predecessors are all binary variables and each predecesser can be thought of as an independent cause of $X_i$. The model can be determined by assessing from the expert a "base" probability for $X_i$ and a single probability for each immediate predecessor of $X_i$. The base probability is the probability that $X_i$ is true when none of its immediate predecessors is true. The model can be estimated by assessing for predecessor $X_j$ the probability that $X_i$ is true when $X_j$ is true and all other predecessors are false. From these values and the assumptions of the noisy-OR model it one can estimate the full set of local conditional distributions for $X_i$. Example 2 shows how the noisy-OR model can be parameterized.

*Example 2: Noisy-OR model*. The variable $X_i$ and its parents $X_j$ ($j \in c(i)$) are binary, taking on values $t_i$ ($t_j$) and $f_i$ ($f_j$). The parameter vector $\theta_i$ consists of one element for each parent variable. An element of $\theta_i$ is denoted by $\theta_{i[j]}$, where the subscript $j$ ranges over the indices $c(j)$ of the conditioning variables. The value $\theta_{i[j]}$ represents the probability that the "inhibitor" corresponding to variable $x_j$ is active (see Pearl, 1989). The variable $X_i$ takes on the value $t_i$ if at least one inhibitor for a parent variable in state $t_j$ is inactive. One may also add a "dummy inhibitor" $\theta_{i[0]}$ which represents a "base probability" that $X_i = t_i$ when none of its parents in is the true state.

$$P(t_i|x_{c(i)}, \theta_i) = 1 - \theta_{i[0]} \prod_{x_j = t_j} \theta_{i[j]}$$

$$P(f_i|x_{c(i)}, \theta_i) = \theta_{i[0]} \prod_{x_j = t_j} \theta_{i[j]} \qquad (3)$$

Other lower dimensional representations of the local conditional distributions are have been discussed in the literature. For example, Srinivas (1992) discusses generalizing the noisy-OR to non-binary variables.

---

[1] I use uppercase letters to represent random variables and lowercase letters to represent specific values for the random variables. Boldface letters represent vectors and standard letters represent scalars.

Asymmetric independencies (e.g., Heckerman, 1990)) can be represented as equality constraints on some of the local conditional probabilities.

In what follows, I assume that an initial value of the parameter vector θ has been specified. This initial specification commonly comes from expert judgment. (Note that the expert need not specify θ values directly; these values can be inferred from probability judgments with which the expert feels comfortable.) The goal of the methodology presented here is to provide guidance to an expert or analyst who wishes to consider the impact of varying the parameter vector. I assume that θ can be varied in an open ball around the value θ. This means that θ does not lie on the boundary of the parameter space and that the $\theta_{i[k]}$ can all be varied independently of each other.

## 2.2 Single Target Probability, Single Scenario

A *scenario* for sensitivity assessment is defined by a *target* variable $X_t$ and an assigned set of values $x_e$ for a subset $X_e$ of variables called *evidence* variables. The goal of sensitivity analysis is to analyze the impact of changes in the parameter vector θ on the probability distribution $P(X_t|x_e,\theta)$. In this section I show how to assess the impach of small changes in a single element $\theta_{s[k]}$ on $P(X_t|x_e,\theta)$. The selected element $\theta_{s[k]}$ is the $k$th element of the parameter vector for the local conditioal distribution for the selected variable $X_s$. Proposition 1 shows how to compute the sensitivity of the probability $P(x_t \mid x_e,\theta_s)$ to changes in the parameter $\theta_{s[k]}$.

**Proposition 1.** The partial derivative of the probability value $P(x_t|x_e,\theta)$ with respect to the parameter value $\theta_{sk}$ is given by

$$\frac{\partial P(x_t|x_e,\theta)}{\partial \theta_{s[k]}} =$$

$$P(x_t|x_e,\theta)\left(E[U_{s[k]}|x_t,x_e] - E[U_{s[k]}|x_e]\right) \quad (4)$$

where

$$U_{s[k]}(X_s,X_{c(s)},\theta_s) = \frac{\partial}{\partial \theta_{s[k]}} \log P(X_s|X_{c(s)},\theta_s)$$

The proof of Proposition 1 is given in the appendix.

The next two propositions apply this result to the two examples described in Section 2.1. To find sensitivity values for the unrestricted models, one needs to compute the function U when the local conditional distribution is defined by (2). This is found by differentiating (2) with respect to $\theta_{i[x_i,x_{c(i)}]}$ and using the chain rule to compute the derivative of the logarithm.

**Proposition 2.** The values of U for the unrestricted node distribution model of Example 1 are given by

$$U_{i[x_i,x_{c(i)}]}(x_i,x_{c(i)},\theta_i) =$$
$$\frac{1}{\sum_{x'_i}\theta_{i[x'_i,x_{c(i)}]}} \left(\frac{1-P(x_i|x_{c(i)},\theta_i)}{P(x_i|x_{c(i)},\theta_i)}\right)$$

$$U_{i[x_i,x_{c(i)}]}(x_i^\circ,x_{c(i)},\theta_i) = -\frac{1}{\sum_{x'_i}\theta_{i[x'_i,x_{c(i)}]}}$$

$$U_{i[x_i,x_{c(i)}]}(x_i^\circ,x_{c(i)}^\circ,\theta_i) = 0 \quad (5)$$

From Proposition 2 it is clear that increasing $\theta_{i[x_i,x_{c(i)}]}$ increases the probability of $x_t$ given $x_{c(i)}$ (this is hardly surprising). Icreasing $\theta_{i[x_i,x_{c(i)}]}$ also decreases the probability of other values of $X_t$ because of the sum-to-1 constraint. The distributions of $X_t$ given other combinations of values of the parent variables are unaffected by changes in $\theta_{i[x_i,x_{c(i)}]}$.

**Proposition 3:** The values of U for the Noisy-OR model of Example 2 are given by

$$U_{i[j]}(t_i,x_{c(i)},\theta_i) = \begin{cases} -\theta_{i[0]} \prod_{\substack{x_{j'}=t_{j'} \\ j' \neq j}} \theta_{i[j']} & x_j = t_j \\ 0 & x_j = f_j \\ -\prod_{x_{j'}=t_{j'}} \theta_{i[j']} & j = 0 \end{cases}$$

$$U_{i[j]}(f_i,x_{c(i)},\theta_i) = \begin{cases} \theta_{i[0]} \prod_{\substack{x_{j'}=t_{j'} \\ j' \neq j}} \theta_{i[j']} & x_j = t_j \\ 0 & x_j = f_j \\ \prod_{x_{j'}=t_{j'}} \theta_{i[j']} & j = 0 \end{cases} \quad (6)$$

Proposition 3 says that changes in the inhibitor probability $\theta_{i[j]}$ change the probability distribution of $X_t$ only when the corresponding parent variable $X_j$ is true. Increases in the inhibitor probability have impact of equal magnitude but opposite sign on the probability of $t_i$ and the probability of $f_i$ conditional on any given combination of values of the parent variables.

## 2.3 Computing Sensitivity Values

The expectations in (4) can be computed by straightforward modification of standard belief network propagation algorithms. I describe how to compute these expectations using the Lauritzen and Spiegelhalter algorithm and variants of logic sampling. Similar modifications are possible with other belief propagation algorithms.

In some cases, the structure of the graph implies that (4) is equal to zero. To determine when (4) is equal to zero, one constructs and auxiliary graph by adding a new parent node $\Theta_i$ to each node $X_i$. This new parent represents the possible values of the parameter vector $\theta_i$. Now, (4) is equal to zero when $\mathbf{X}_e$ $d$-separates $\Theta_s$ from $\Theta_t$. In addition, as evidenced by Equations (2) and (3), specific local models imply additional conditions under which the sensitivities of $\theta_{t[k]}$ are zero.

The Lauritzen and Spiegelhalter algorithm and its variants transform the network into a tree of cliques, where each clique consists of a subset of variables in the network and the cliques satisfy the running intersection property (Neapolitan, 1990). The set consisting of a node $X_j$ and its parents $\mathbf{X}_{c(j)}$ must belong to at least one clique. The belief propagation algorithm computes and stores with each clique a *potential* for each combination of values of each node in the clique. The clique potential function is proportional to the joint conditional probability distribution over the nodes in the clique given the values of the evidence nodes.

To compute sensitivity values for $X_t = x_t$, first declare as evidence $X_t = x_t$ and $\mathbf{X}_e = \mathbf{x}_e$ and run the belief propagation algorithm. As noted above, for each $s$, one of the cliques must contain $X_s$ and all its parents $\mathbf{X}_{c(s)}$. Marginalizing the clique potential over all nodes in this clique other than $(X_s, \mathbf{X}_{c(s)})$ yields a function which can be normalized to obtain the conditional probability distribution $P(X_s, \mathbf{X}_{c(s)}|x_t, \mathbf{x}_e, \theta)$. Now, use this joint distribution to compute the second expectation in (4):

$$E[U_{s[k]}|x_t, \mathbf{x}_e] = \qquad (7)$$

$$\sum_{(x_s, x_{c(s)})} U_{s[k]}(x_s, x_{c(s)}, \theta_s) P(x_s, x_{c(s)}|x_t, \mathbf{x}_e, \theta)$$

In this manner, compute $E[U_{s[k]}]$ given each value $x_t$ of $X_t$. The algorithm must now be run again without conditioning on $X_t$ (but still conditioning on $\mathbf{X}_e = \mathbf{x}_e$). The first expectation in (4) is now computed as follows:

$$E[U_{s[k]}|\mathbf{x}_e] = \sum_{x_t} E[U_{s[k]}|x_t, \mathbf{x}_e] P(x_t|\mathbf{x}_e, \theta) \qquad (8)$$

The values (7) and (8) may be substituted into (4) to yield the sensitivity value.

Approximating (4) using Monte Carlo simulation is also straightforward. For each node $X_s$ for which sensitivities are to be calculated, define arrays $A_{s[k]}$ and $B_{s[k]}$, each with one element for each possible value of $X_t$. Initialize the array values to zero. Each iteration of the algorithm yields a realization $x$ of all nodes in the network and a sampling weight $w$.[2] Compute $U_{s[k]}(x_s, x_{c(s)}, \theta_s)$ and increment $A_{s[k]}(x_t)$ by its value. Increment $B_{s[k]}(x_t)$ by $w$.

After the simulation is run, compute the estimates

$$\hat{E}[U_{s[k]}|x_t, \mathbf{x}_e] = \frac{A_{s[k]}(x_t)}{B_{s[k]}(x_t)} \quad \text{and}$$

$$\hat{E}[U_{s[k]}|\mathbf{x}_e] = \frac{\sum_{x_t} A_{s[k]}(x_t)}{\sum_{x_t} B_{s[k]}(x_t)}. \qquad (9)$$

These estimates can be plugged into (4) to estimate the desired sensitivity values.

## 3. EXAMPLE

The graph of Figure 1 is taken from Neapolitan (1990), from an example originally due to Lauritzen and Spiegelhalter. The local probability distributions for this example are given in Table 1. Sensitivity values for this network were computed as described in Section 2.3 using the Lauritzen and Spiegelhalter algorithm as implemented in IDEAL (Srinivas and Breese, 1992), using the unrestricted node model of Example 1. The evidence nodes were A=$t_A$ and H=$t_H$ (dyspnea observed in a patient who had been to Asia). Sensitivities were computed for target value B=$t_B$ (patient has tuberculosis). Table 2 summarizes the sensitivity information for each node's local probability distribution. The value shown is the largest value of $|U_{s[x_s, x_{c(s)}]}|$ for the node. Variable C is not shown. Because C is defined as a deterministic function of the values of its parent variables B and E, so sensitivities for C are not of interest.

The local distribution to which the target probability $P(B=t_B|A=t_A,H=t_H)$ is most sensitive is the distribution of B given A=$t_A$. The maximum sensitivity value for the local distribution of B is about 1.6. This value corresponds to the distribution of B when A=$t_A$ (the sensitivity for A=$f_A$ is zero because the conditional distribution of B given $f_A$ is irrelevant given the scenario in which A=$t_A$). This large value is not surprising because the evidence variable A is a direct predecessor of B. One certainly expects P(B|A,E) to be highly sensitive to P(B|A).

The largest value for this node is greater by a factor of 18 than maximum value for node H (dyspnea), another local distribution for an evidence node. The local distributions for E (lung cancer) and G (bronchitis), both competing explanations for the finding of dyspnea, have maximum sensitivities about half that for node H. The node F (smoking), which affects the prior probabilities of both E and G, has about half the maximum sensitivity value of either of these nodes. Finally, all sensitivities for A and D are zero. In the absense of any observations, the local conditional distribution of D has no effect on the posterior probability of B. The prior probability of A is irrelevant once the value of A becomes known.

---

[2]The sampling weight adjusts for the effect of efficiency improving modifications such as likelihood weighting and importance sampling. A weight of zero is given to any observation for which $\mathbf{X}_e \neq \mathbf{x}_e$.

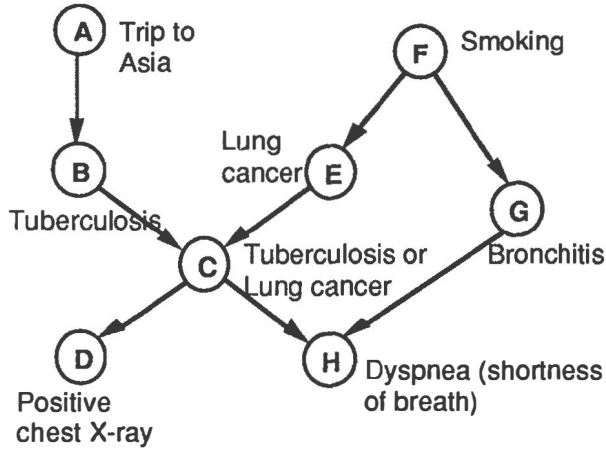

Figure 1: Example Network

| | | | | |
|---|---|---|---|---|
| $P(t_A)$ | = .01 | | | |
| $P(t_B\|t_A)$ | = .05 | $P(t_A\|f_B)$ | = | .01 |
| $P(t_C\|t_Bt_E)$ | = 1 | $P(t_C\|t_Bf_E)$ | = | 1 |
| $P(t_C\|f_Bt_E)$ | = 1 | $P(t_C\|f_Bf_E)$ | = | 0 |
| $P(t_D\|t_C)$ | = .98 | $P(t_D\|f_C)$ | = | .05 |
| $P(t_E\|t_F)$ | = .1 | $P(t_E\|f_F)$ | = | .01 |
| $P(t_F)$ | = .5 | | | |
| $P(t_G\|t_F)$ | = .6 | $P(t_G\|f_F)$ | = | .3 |
| $P(t_H\|t_Ct_G)$ | = .9 | $P(t_H\|t_Cf_G)$ | = | .7 |
| $P(t_H\|f_Ct_G)$ | = .8 | $P(t_H\|f_Cf_G)$ | = | .1 |

Table 1: Probabilities for Dyspnea Example

| Node | | Maximum Absolute Sensitivity |
|---|---|---|
| A | Asia visit | 0 |
| B | Tuberculosis | 1.6 |
| D | Positive X-ray | 0 |
| E | Lung Cancer | 0.041 |
| F | Smoker | 0.019 |
| G | Bronchitis | 0.038 |
| H | Dyspnea | 0.088 |

Table 2: Sensitivities for Dyspnea Example
Evidence $A=t_A, H=t_H$; Target $B=t_B$

## 4. INCORPORATING DIRECT ESTIMATES OF TARGET DISTRIBUTIONS

### 4.1 Adjusting to Fit a Directly Assessed Distribution

Suppose the expert provides not only the local conditional probability distributions $P(x_s|x_{c(s)}, \theta_s)$ but also direct assessments of target probability distributions for a set of scenarios. I consider here how to use these direct assessments in estimating model parameters.

Begin with an initial model $P(x|\theta)$. Suppose the expert is given a scenario $X_e=x_e$ and assesses directly the distribution of a target variable $P^*(x_t|x_e)$. In general this will be different from the model distribution $P(x_t|x_e,\theta)$. Suppose the expert wishes guidance on how to change the model to bring $P(x_t|x_e,\theta)$ closer to $P^*(x_t|x_e)$.

A way to measure distance between $P(x_t|x_e,\theta)$ and $P^*(x_t|x_e)$ is to use a *proper scoring rule*. A scoring rule assigns a score $s(x_t, P)$ if outcome $x_t$ occurs and the probability distribution P was assessed. If $P^*$ is the correct distribution, the expected score is

$$d(P,P^*) = E_{P^*}[s(X_t,P^*)-s(X_t,P)]$$
$$= \sum_{x_t}(s(x_t,P^*)-s(x_t,P))P^*(x_t|x_e) \quad (10)$$

A scoring rule is proper if (10) is always positive (i.e., one maximizes one's expected score by assessing the correct distribution). I make the assumption that (10) can be written as

$$d(P,P^*) = \sum_{x_t} h(P(x_t|x_e,\theta), P^*(x_t|x_e))P^*(x_t|x_e). \quad (11)$$

This is the case for two of the most common scoring rules, the quadratic and logarithmic rules as defined in Lindley (1982). For these two rules, $h(P,P^*)$ is given by

$$h_{\log}(P,P^*) = \log(P) - \log(P^*) \quad \text{and} \quad (12)$$

$$h_{\text{quad}}(P,P^*) = P^*(1-P^*) + (P-P^*)^2. \quad (13)$$

respectively.

Using the results of the previous section, the partial derivative of (10) with respect to $\theta_{s[k]}$ can be computed:

$$\frac{\partial}{\partial \theta_{s[k]}} d(P,P^*) = \quad (14)$$

$$\sum_{x_t} \left(\frac{\partial}{\partial P} h(P,P^*)\right)\left(\frac{\partial}{\partial \theta_{s[k]}} P(x_t|x_e,\theta)\right) P^*(x_t|x_e)$$

which is straightforward to compute from (4) and the distributions P and P*.

It is interesting to note that for the logarithmic scoring rule (14) reduces to

$$\frac{\partial}{\partial \theta_{s[k]}} d(P,P^*) =$$

$$\sum_{x_t} \left(E[U_{s[k]}|x_t,x_e] - E[U_{s[k]}|x_e]\right) P^*(x_t|x_e)$$

$$= E^*[U_{s[k]}|x_e] - E[U_{s[k]}|x_e] \quad (15)$$

The first expression on the right side of (15) denotes the expectation of $U_{S[k]}$ taken under the distribution $Q(X) = P^*(X_t)P(X|X_t) = P(X)(P^*(X_t)/P(X_t))$. That is, (15) is the difference in the expectation of $U_{S[k]}$ under two distributions, one in which the distribution of $x_t$ is set equal to the holistically assessed distribution and the other equal to the model distribution.

If holistic assessments are made for a number of scenarios, (15) can be computed and examined separately for each scenario. Alternatively, an aggregate goodness-of-fit measure can be computed by summing values of (10) for different scenarios (the sum can be weighted by importance of the assessment or by a measure of how sure the expert is of the judgment). The appropriate derivative then is the corresponding (perhaps weighted) sum of values of (15).

### 4.2 Automating Best-Fit Assessments

The methods presented here can also be used to compute a best-fit model (under one of the scoring rules presented in Section 3.1) given a set of judgments (holistic and local) from an expert. A simple gradient descent method can be defined as follows.

1. Initialize the network probabilities.
2. Select a scenario. (A scenario is a conditional probability distribution assessed by the expert. A local conditional distribution counts as a scenario. Scenario selection may be random or may cycle through the scenarios in some fixed order.)
3. Compute (15) for all relevant nodes. (D-separation can be used to eliminate some computations. If the scenario is an assessment of a local conditional distributions, only the distribution for that node need be considered.)
4. Change all relevant $\theta_{S[k]}$ by an amount proportional to (15).
5. Cycle through Steps 2 through 4 until a convergence criterion is met.

This gradient descent approach is employed by common neural network learning algorithms (e.g., backpropagation and Boltzmann machine learning; see Laskey, 1990).

This algorithm may stop at a local optimum (the objective function is generally not convex in the parameters). If all local probability assessments are available, they determine a consistent global model which may make a good starting value for θ. It may be desirable to restart the algorithm from different starting values. Cycling through scenarios in random order introduces a random element to the algorithm, which may help prevent its becoming stuck in local optima.

Of course, it is always a good idea to identify and set aside for special examination any outliers, or assessments for which the estimated model fits very poorly.

## 5. DISCUSSION

This paper describes a method for computing the sensitivity of a target probability or a target distribution to changes in network parameters. Sensitivity values can be computed one scenario (instantitation of evidence variables) at a time, or sensitivities of an aggregate goodness-of-fit measure for multiple scenarios can be computed. The method can be adapted for automated fitting of a best-fitting model to a set of holistic and local judgments.

This paper considered the problem of adjusting the parameter values in a model with fixed structure. Another important part of the knowledge elicitation process is changing the structure of a model to better fit the expert's judgments. Sensitivity values can also be used to suggest links to add. The absence of a link in the network can be viewed as the assignment of a zero value to a log-linear interaction term. For each link one wishes to consider adding to the network, one can compute a sensitivity value for this parameter. If the sensitivity value is large, there is a large improvement in model fit by adding the extra link.

### References

Buntine, W. (1991) Theory Refinement on Bayesian Networks. In D'Ambrosio, Smets and Bonissone (eds) *Uncertainty in Artificial Intelligence: Proceedings of the Seventh Conference*. San Mateo, CA: Morgan Kaufman.

Cooper, G. and Herskovits, E. (1990) A Bayesian Method for Constructing Bayesian Belief Networks from Databases. In D'Ambrosio, Smets and Bonissone (eds) *Uncertainty in Artificial Intelligence: Proceedings of the Seventh Conference*. San Mateo, CA: Morgan Kaufman.

Heckerman, D. (1990) *Probabilistic Similarity Networks*. Ph.D. Thesis, Departments of Computer Science and Medicine, Stanford University.

Laskey, K.B. (1990) Adapting Connectionist Learning to Bayes Networks, *International Journal of Approximate Reasoning*, 4, 261-282.

Lindley, D. V. (1982). Scoring Rules and the Inevitability of Probability. *International Statistical Review*, 50, 1-26.

Neapolitan, R.E. (1990) *Probabilistic Reasoning in Expert Systems: Theory and Applications*. New York: Wiley.

Pearl, J. (1989) *Probabilistic Reasoning in Intelligent Systems*. San Mateo, CA: Morgan Kaufman.

Spiegelhalter, D. and Lauritzen (1990). Sequential Updating of Conditional Probabilities on Directed Graphical Structures, *Networks* 20, 579-605.

Srinivas, S. (1992) *Generalizing the Noisy Or concept to non-binary variabes*. Technical Report No. 79, Rockwell International Science Center.

Srinivas, S. and Breese, J. (1992) *IDEAL: Influence Diagram Evaluation and Analysis in Lisp: Documentation and Users Guide*. Technical Memorandum No. 23, Palo Alto, CA: Rockwell International Science Center.

## APPENDIX: PROOFS OF RESULTS

The proof of Proposition 1 requires the following lemma.

**Lemma 1.** The partial derivative of the unconditional probability value $P(x|\theta)$ with respect to the parameter value $\theta_{s[k]}$ is given by:

$$\frac{\partial}{\partial \theta_{s[k]}} P(x|\theta) = P(x|\theta) U_{s[k]}(x_s, x_{c(s)}, \theta_s) \quad (A-1)$$

**Proof of Lemma 1:**

$$\frac{\partial}{\partial \theta_{s[k]}} P(x|\theta) =$$

$$\prod_{i \neq s} P(x_i | x_{c(i)}, \theta_i) \left( \frac{\partial}{\partial \theta_{s[k]}} P(x_s | x_{c(s)}, \theta_s) \right)$$

$$= P(x|\theta) \frac{\frac{\partial}{\partial \theta_{s[k]}} P(x_s | x_{c(s)}, \theta_s)}{P(x_s | x_{c(s)}, \theta_s)}$$

$$= P(x|\theta) \frac{\partial}{\partial \theta_{s[k]}} \log P(x_s | x_{c(s)}, \theta_s) . \quad \blacksquare$$

**Proof of Proposition 1:**

$$\frac{\partial P(x_t | x_e, \theta)}{\partial \theta_{s[k]}} =$$

$$\frac{\partial}{\partial \theta_{s[k]}} \left( \frac{\sum_{x_s, x_u} P(x_t, x_s, x_u, x_e | \theta)}{\sum_{x'_t, x_s, x_u} P(x'_t, x_s, x_u, x_e | \theta)} \right)$$

$$= \frac{\sum_{x_s, x_u} P(x_t, x_s, x_u, x_e | \theta) U_{s[k]}(x_s, x_{c(s)}, \theta_s)}{\sum_{x'_t, x_s, x_u} P(x'_t, x_s, x_u, x_e | \theta)}$$

$$- \frac{\left( \sum_{x_s, x_u} P(x_t, x_s, x_u, x_e | \theta) \right) \times \left( \sum_{x'_t, x_s, x_u} P(x'_t, x_s, x_u, x_e | \theta) U_{s[k]}(x_s, x_{c(s)}, \theta_s) \right)}{\left( \sum_{x'_t, x_s, x_u} P(x'_t, x_s, x_u, x_e | \theta) \right)^2}$$

$$= \sum_{x_s, x_u} P(x_t, x_s, x_u | x_e, \theta) U_{s[k]}(x_s, x_{c(s)}, \theta_s)$$

$$- \left( \sum_{x_s, x_u} P(x_t, x_s, x_u | x_e, \theta) \right) \times \left( \sum_{x'_t, x_s, x_u} P(x'_t, x_s, x_u | x_e, \theta) U_{s[k]}(x_s, x_{c(s)}, \theta_s) \right)$$

$$= P(x_t | x_e, \theta) \times \sum_{x_s, x_u} P(x_s, x_u | x_t, x_e, \theta) U_{s[k]}(x_s, x_{c(s)}, \theta_s)$$

$$- P(x_t | x_e, \theta) \times \left( \sum_{x'_t, x_s, x_u} P(x'_t, x_s, x_u | x_e, \theta) U_{s[k]}(x_s, x_{c(s)}, \theta_s) \right)$$

$$= P(x_t | x_e, \theta) \Big( E\big[ U_{s[k]}(x_s, x_{c(s)}, \theta_s) \big| x_t, x_e \big]$$

$$- E\big[ U_{s[k]}(x_s, x_{c(s)}, \theta_s) \big| x_e \big] \Big). \quad \blacksquare$$